\pdfoutput=1
\documentclass[10pt,twocolumn,letterpaper]{article}

\usepackage[pagenumbers]{cvpr} %

\usepackage{graphicx}
\usepackage{amsmath}
\usepackage{amssymb}
\usepackage{booktabs}
\usepackage{lipsum}

\usepackage{array}
\usepackage{times}
\usepackage{epsfig}
\usepackage{graphicx}
\usepackage{float}
\usepackage{wrapfig}
\usepackage{amsmath,amssymb,amsthm}
\usepackage{algorithm,algorithmicx,algpseudocode}
\usepackage{bm,xspace}
\usepackage{comment}
\usepackage{multirow}
\usepackage{balance}
\usepackage{url}
\usepackage{booktabs}
\usepackage{etoolbox,siunitx}
\usepackage{calc}
\usepackage{pifont,hologo}
\usepackage{color}
\usepackage{adjustbox}
\usepackage[normalem]{ulem}  %
\usepackage[table]{xcolor}
\usepackage[pagebackref,breaklinks,colorlinks,linkcolor=red,citecolor=blue]{hyperref}
\usepackage[accsupp]{axessibility}  %

\newcommand{\authorskip}{\hspace{4mm}}

\definecolor{blue}{HTML}{0055cc}
\definecolor{red}{HTML}{cc1100}
\definecolor{orange}{HTML}{cc7700}
\definecolor{gray}{HTML}{efefef}
\definecolor{darkgreen}{rgb}{0.13, 0.55, 0.13}
\definecolor{darkgray}{HTML}{757575}

\newcommand{\figref}[1]{Figure~\ref{#1}}
\newcommand{\tabref}[1]{Table~\ref{#1}}
\newcommand{\secref}[1]{Section~\ref{#1}}
\renewcommand{\eqref}[1]{Eq.~\ref{#1}}

\newcommand{\marktext}[2]{\adjustbox{bgcolor=#1}{\strut #2}}

\newcolumntype{x}[1]{>{\centering\arraybackslash}p{#1}}
\newcolumntype{y}[1]{>{\raggedright\arraybackslash}p{#1}}
\newcolumntype{z}[1]{>{\raggedleft\arraybackslash}p{#1}}
\newcommand{\tablestyle}[2]{\setlength{\tabcolsep}{#1}\renewcommand{\arraystretch}{#2}\centering\footnotesize}

\DeclareMathSymbol{@}{\mathord}{letters}{"3B}

\newcommand\mypara[1]{\vspace{0mm}\noindent\textbf{#1}}

\makeatletter
\DeclareRobustCommand\onedot{\futurelet\@let@token\@onedot}
\def\@onedot{\ifx\@let@token.\else.\null\fi\xspace}
\def\eg{e.g\onedot}

 \def\etal{et al\onedot}

\newcommand*{\Rom}[1]{\expandafter\@slowromancap\romannumeral #1@}
\newcommand*{\rom}[1]{\expandafter\romannumeral #1}

\def\1{\bm{1}}

\def\vc{{\bm{c}}}

\def\vf{{\bm{f}}}

\def\vp{{\bm{p}}}

\def\vt{{\bm{t}}}

\def\vx{{\bm{x}}}

\def\mC{{\bm{C}}}

\def\mF{{\bm{F}}}

\def\mM{{\bm{M}}}

\def\mP{{\bm{P}}}

\def\mS{{\bm{S}}}
\def\mT{{\bm{T}}}

\def\mX{{\bm{X}}}

\DeclareMathAlphabet{\mathsfit}{\encodingdefault}{\sfdefault}{m}{sl}
\SetMathAlphabet{\mathsfit}{bold}{\encodingdefault}{\sfdefault}{bx}{n}

\def\sP{{\mathcal{P}}}

\usepackage[capitalize]{cleveref}
\crefname{section}{Sec.}{Secs.}
\Crefname{section}{Section}{Sections}
\Crefname{table}{Table}{Tables}
\crefname{table}{Tab.}{Tabs.}

\def\cvprPaperID{3221} %
\def\confName{CVPR}
\def\confYear{2023}

\begin{document}

\title{Masked Scene Contrast: A Scalable Framework for \\ Unsupervised 3D Representation Learning}

\author{
Xiaoyang Wu \authorskip Xin Wen \authorskip Xihui Liu \authorskip Hengshuang Zhao\thanks{Corresponding Author. Email: {\tt\footnotesize \href{mailto:hszhao@cs.hku.hk}{hszhao@cs.hku.hk}}} \\
The University of Hong Kong\\
{\tt\small \url{https://github.com/Pointcept/Pointcept}}
}
\maketitle

\begin{abstract}
  As a pioneering work, PointContrast conducts unsupervised 3D representation learning via leveraging contrastive learning over raw RGB-D frames and proves its effectiveness on various downstream tasks. However, the trend of large-scale unsupervised learning in 3D has yet to emerge due to two stumbling blocks: the inefficiency of matching RGB-D frames as contrastive views and the annoying mode collapse phenomenon mentioned in previous works. Turning the two stumbling blocks into empirical stepping stones, we first propose an efficient and effective contrastive learning framework, which generates contrastive views directly on scene-level point clouds by a well-curated data augmentation pipeline and a practical view mixing strategy. Second, we introduce reconstructive learning on the contrastive learning framework with an exquisite design of contrastive cross masks, which targets the reconstruction of point color and surfel normal. Our Masked Scene Contrast~(MSC) framework is capable of extracting comprehensive 3D representations more efficiently and effectively. It accelerates the pre-training procedure by at least 3$\times$ and still achieves an uncompromised performance compared with previous work. Besides, MSC also enables large-scale 3D pre-training across multiple datasets, which further boosts the performance and achieves state-of-the-art fine-tuning results on several downstream tasks, \eg, 75.5\% mIoU on ScanNet semantic segmentation validation set.
\end{abstract}

\begin{figure}[t]\centering
\includegraphics[width=0.87\linewidth]{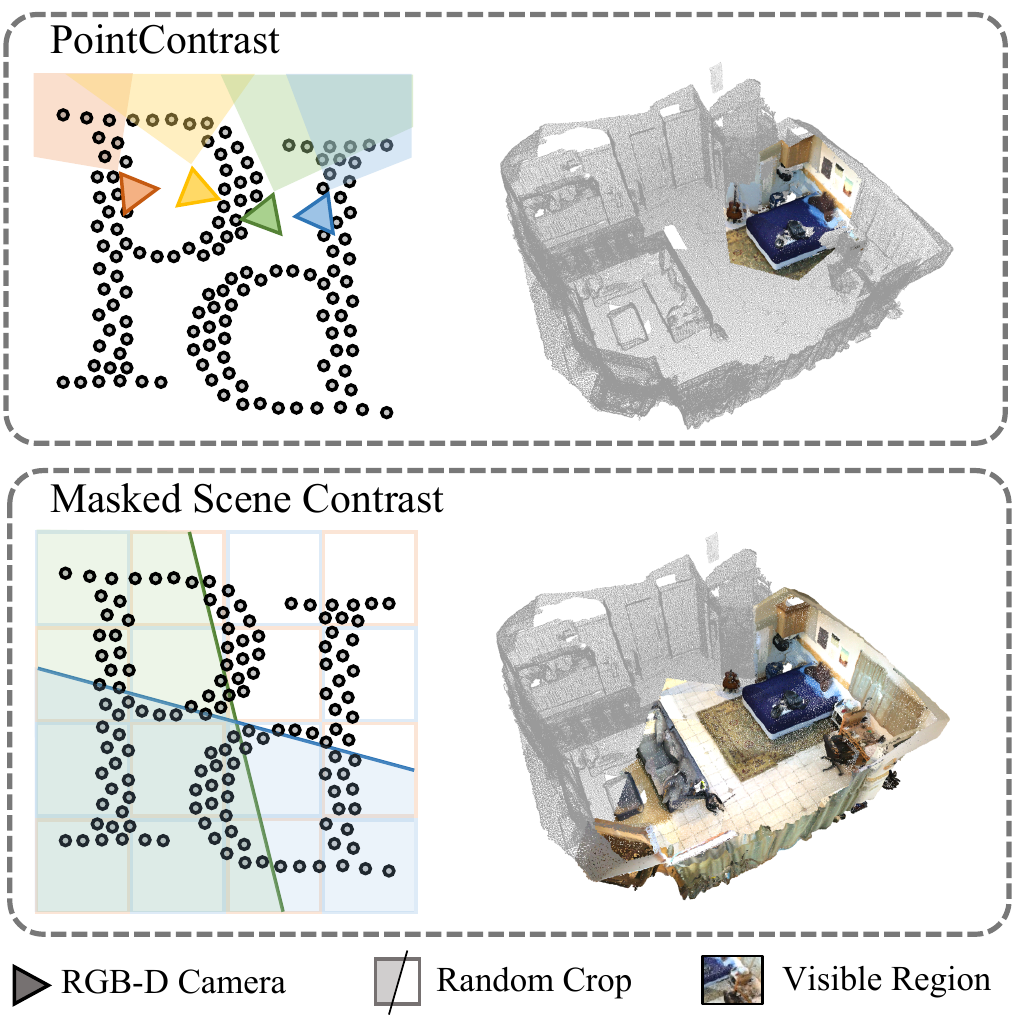}
\vspace{-4mm}
\caption{Comparison of unsupervised 3D representation learning. The previous method~\cite{xie2020pointcontrast} (top) relies on raw RGB-D frames with restricted views for contrastive learning, resulting in low efficiency and inferior versatility. Our approach (bottom) directly operates on scene-level views with contrastive learning and masked point modeling, leading to high efficiency and superior generality, further enabling large-scale pre-training across multiple datasets.}
\label{fig:teaser}
\vspace{-6mm}
\end{figure}

\section{Introduction}
Unsupervised visual representation learning aims at learning visual representations from vast amounts of unlabeled data. The learned representations are proved to be beneficial for various downstream tasks like segmentation and detection. It has attracted lots of attention and achieved remarkable progress in 2D image understanding, exceeding the upper bound of human supervision~\cite{grill2020bootstrap,henaff2020data}.

Despite the impressive success of unsupervised visual representation learning in 2D, it is underexplored in 3D. Modern 3D scene understanding algorithms~\cite{choy20194d, wu2022point} are focused on supervised learning, where models are trained directly from scratch on targeted datasets and tasks. Well-pre-trained visual representations can undoubtedly boost the performance of these algorithms and are currently in urgent demand. Recent work PointContrast~\cite{xie2020pointcontrast} conducts a preliminary exploration in 3D unsupervised learning. However, it is limited to raw RGB-D frames with an inefficient learning paradigm, which is not scalable and applicable to large-scale unsupervised learning. To address this essential and inevitable challenge, we focus on building a scalable framework for large-scale 3D unsupervised learning.

One technical stumbling block towards large-scale pre-training is the inefficient learning strategy introduced by matching RGB-D frames as contrastive views. PointContrast~\cite{xie2020pointcontrast} opens the door to pre-training on real indoor scene datasets and proposes frame matching to generate contrastive views with natural camera views, as in Figure~\ref{fig:teaser} top. However, frame matching is inefficient since duplicated encoding exists for matched frames, resulting in limited scene diversity in batch training and optimization. Meanwhile, not all of the 3D scene data contains raw RGB-D frames, leading to failure deployments of the algorithm. Inspired by the great success of SimCLR~\cite{chen2020simple}, we investigate generating strong contrastive views by directly applying a series of well-curated data augmentations to scene-level point clouds, eliminating the dependence on raw RGB-D frames, as in Figure~\ref{fig:teaser} bottom. Combined with an effective mechanism that mixes up query views, our contrastive learning design accelerates the pre-training procedure by 4.4$\times$ and achieves superior performance with purely point cloud data, compared to PointContast with raw data. The superior design also enables large-scale pre-training across multiple datasets like ScanNet~\cite{dai2017scannet} and ArkitScenes~\cite{dehghan2021arkitscenes}.

Another obstacle is the mode collapse phenomenon that occurs when scaling up the optimization iterations. We owe the culprit for this circumstance to the insufficient difficulty of unsupervised learning tasks. To further tackle the mode collapse challenge in unsupervised learning and scale up the optimization iterations, inspired by recent masked autoencoders~\cite{he2022masked, xie2022simmim}, we construct a masked point modeling paradigm where both point color reconstruction objective and surfel normal reconstruction objective are proposed to recover the masked color and geometric information of the point cloud respectively. We incorporate the mask point modeling strategy into our contrastive learning framework via an exquisite design of contrastive cross masks, leading towards a scalable unsupervised 3D representation learning framework, namely Masked Scene Contrast~(MSC).

Our framework is efficient, effective, and scalable. We conduct extensive experimental evaluations to validate its capability. On the popular point cloud dataset ScanNet, our algorithm accelerates the pre-training procedure by more than 3$\times$, and achieves better performance on downstream tasks, when compared to the previous representative PointContrast. Besides, our method also enables large-scale 3D pre-training across multiple datasets, leading to state-of-the-art fine-tuning results on several downstream tasks, e.g. 75.5\% mIoU on ScanNet semantic segmentation validation set. In conclusion, our work opens up new possibilities for large-scale unsupervised 3D representation learning.

\section{Related Work}
\mypara{2D Image contrastive learning.}
Based on the instance discrimination~\cite{dosovitskiy2015discriminative} pretext task, and combined with the contrastive learning~\cite{oord2019representation,hjelm2018learning} paradigm, modern variants of 2D image contrastive representation learning have shown strong abilities in learning transferable visual representations~\cite{wu2018unsupervised,chen2020simple,he2020momentum}. With the learning objective built on the similarity between randomly augmented image views, this line of work strongly depends on a large batch size~\cite{chen2020simple,he2020momentum} and a finely designed data augmentation pipeline~\cite{chen2020simple,tian2020makes,grill2020bootstrap} to achieve better performance. We find these two points also hold true for 3D contrastive learning.

\mypara{2D Image reconstructive learning.}
In 2D unsupervised learning, there is also a recent trend of switching the pretext task from instance discrimination~\cite{chen2020simple,he2020momentum,grill2020bootstrap,caron2021emerging,bardes2022vicreg} to masked image modeling~\cite{bao2022beit,he2022masked,xie2022simmim,zhou2022image,wei2022masked}. Based on a denoising autoencoder~\cite{vincent2010stacked}-style architecture, the task is to reconstruct the RGB value~\cite{he2022masked,xie2022simmim}, discrete token~\cite{bao2022beit,zhou2022image}, or feature~\cite{wei2022masked} of masked pixels. This line of work has shown strong potential in learning representations on large-scale datasets and is less prone to model collapsing like instance-discrimination-based methods (\eg, contrastive learning). When combined with contrastive learning~\cite{el2021large,assran2022masked,wu2022extreme}, the performance can be further boosted, yet with less dependence on data scale~\cite{el2021large}.

\begin{figure*}[!t]
    \raggedleft
    \subfloat[
        \textbf{Frame matching~\cite{xie2020pointcontrast, hou2021exploring}.} 1. Extract raw RGB-D frames and camera positions from raw data. 2. Project each 2D frame into 3D space produces frame-level point cloud views. 3. Calculate pairwise overlapping rates among each frame of a single view and select pairs with overlapping rates larger than 30\% as pairs of contrastive views. 
        \label{fig:frame_matching}
    ]{
        \includegraphics[width=0.485\textwidth]{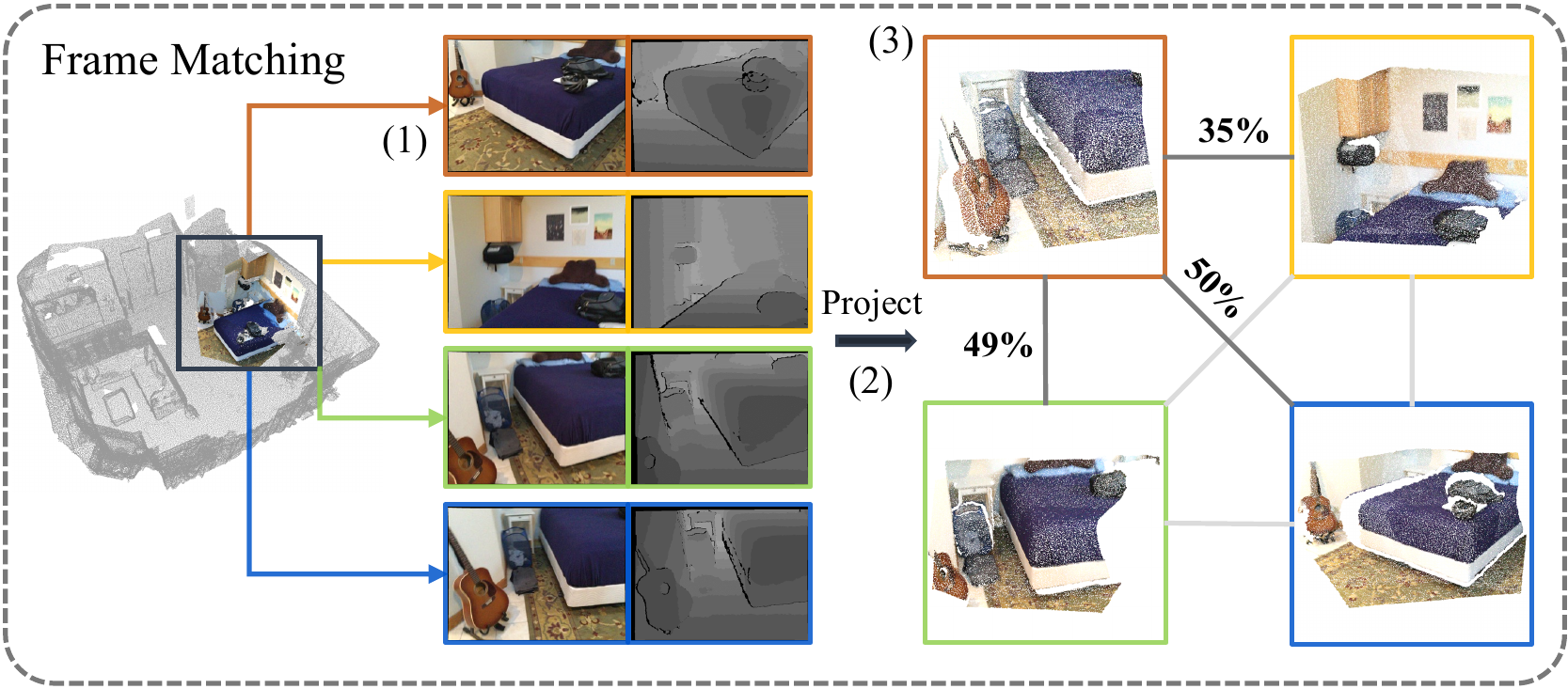}
    }
    \hspace{0.9mm}
    \subfloat[
        \textbf{Scene augmentation~(ours).} 1. Apply spatial augmentations containing rotation, flipping, and scaling. 2. Apply photometric augmentations containing brightness, contrast, saturation, hue, and gaussian noise jittering. 3. Generate and apply contrastive cross masks to the two views after sampling augmentations containing cropping and voxelization.
        \label{fig:scene_augmentation}
    ]{
        \includegraphics[width=0.485\textwidth]{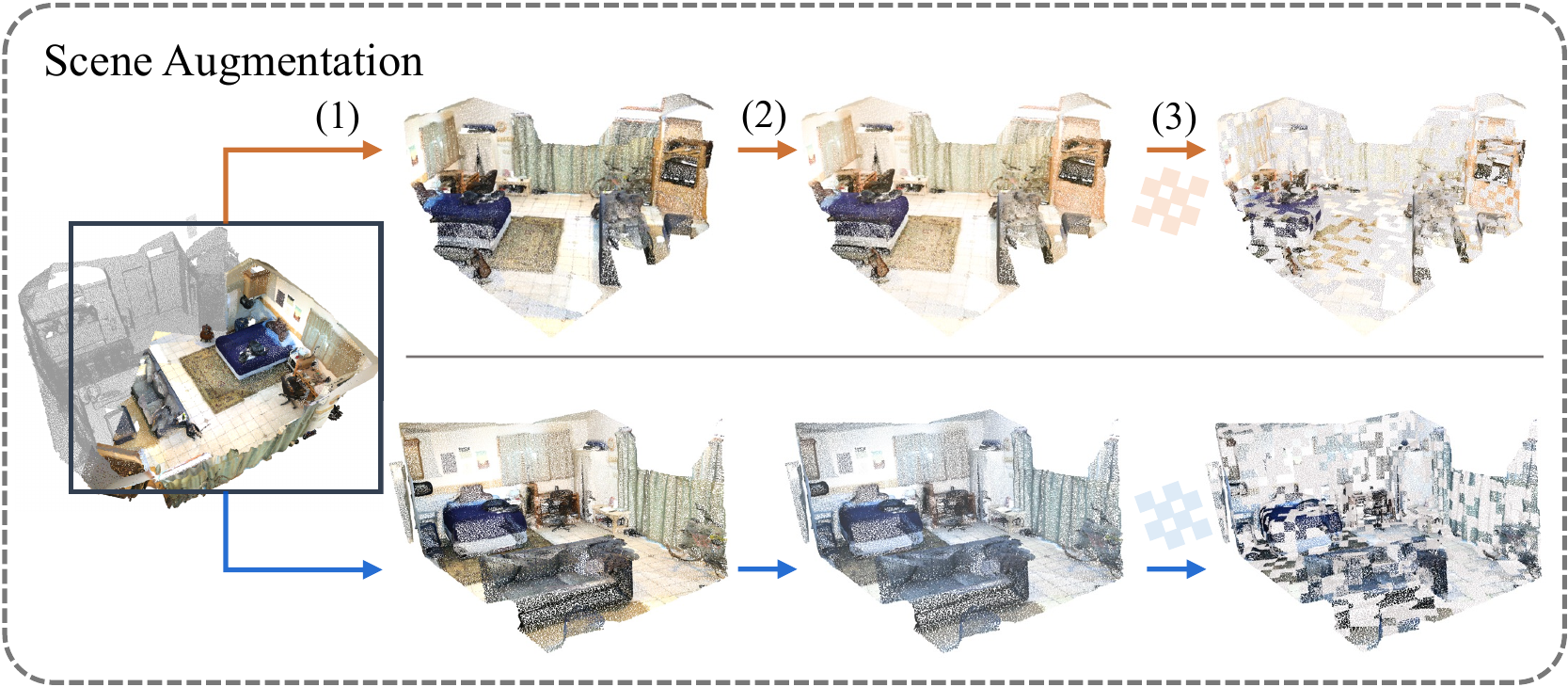}
    }
    \vspace{-1mm}
    \caption{\textbf{View generation.} Compared with frame matching (FM), our scene augmentation (SA) is efficient and effective. 1. SA can end-to-end produce contrastive views on the original point cloud with ignorable latency, while FM requires preprocessing devouring enormous storage resources (\eg additional 1.5 TB storage for ScanNet) in step 2 and pairwise matching is time-consuming. 2. SA produces scene-level views, while FM can only produce frame-level views containing limited information. Benefiting from advanced photometric augmentations, SA has the capacity to simulate the same scene under different lighting.}
    \label{fig:view_generation}
    \vspace{-2mm}
\end{figure*}

\mypara{3D Scene understanding.}
The deep neural architectures for understanding 3D scenes can be roughly categorized into three paradigms according to the way they model point clouds: projection-based, voxel-based, and point-based.
Projection-based works project 3D points into various image planes and adopt 2D CNN-based backbones to extract features\cite{su15mvcnn,li2016vehicle,chen2017multi,lang2019pointpillars}.
On the other hand, the voxel-based stream transforms point clouds into regular voxel representations to operate 3D convolutions~\cite{maturana2015voxnet,song2017semantic}.
Their efficiency is then improved thanks to sparse convolution~\cite{graham20183d,choy20194d,chu2022twist}.
Point-based methods, in contrast, directly operate on the point cloud~\cite{qi2017pointnet,qi2017pointnet++,zhao2019pointweb,thomas2019kpconv,chu2021icm}, and see a recent transition towards transformer-based architectures~\cite{guo2021pct,zhao2021point,wu2022point}.
Following~\cite{xie2020pointcontrast}, we mainly pre-train on the voxel-based method SparseUNet~\cite{choy20194d} implemented with SpConv~\cite{spconv2022}.

\mypara{3D Representation learning.}
Unlike the 2D counterparts, where large-scale unsupervised pre-training has been a common choice for facilitating downstream tasks~\cite{caron2021emerging}, 3D representation learning is still not mature, and most works still train from scratch on the target data directly~\cite{hou2021exploring}.
While earlier works in 3D representation learning simply build on a single object~\cite{wang2019deep,hassani2019unsupervised,sauder2019self,sanghi2020info3d}, recent works start to train on scene-centric point clouds~\cite{xie2020pointcontrast,hou2021exploring}.
However, unlike in 2D that scene-centric representation learning has been well-studied~\cite{liu2020self,xie2021unsupervised,wen2022self}, the pre-training on 3D scenes, which relies on raw frame data~\cite{xie2020pointcontrast,hou2021exploring}, still faces inefficiency issues and finds it hard to scale up to larger scale datasets.
In contrast, we explore directly learning at the scene level, which shows significantly higher efficiency in processing scene data, and opens the possibility for pre-training with larger-scale point clouds, for the first time ever.

\section{Pilot Study}
\label{sec:pilot}
This section analyzes the two main obstacles towards large-scale pre-training with point clouds. Our proposed design is based on the conclusion of the pilot study.

\mypara{Is matching RGB-D frames a good choice?} \\
As a seminal work in 3D representation learning, PointContrast~\cite{xie2020pointcontrast} first enables pre-training in real-world indoor scenes with matched raw RGB-D frames as contrastive views. A visualization of the frame matching procedure is illustrated in \figref{fig:frame_matching}. This protocol seems natural for indoor scenes since point clouds of indoor scenes are usually derived from RGB-D videos~\cite{armeni2016s3dis,dai2017scannet,dehghan2021arkitscenes}, where the raw frames are extracted from. However, this framework has multiple drawbacks that can hinder the scalability of training:
\begin{itemize}
    \vspace{-2mm}
    \item \textit{Redundant frame encoding.} The pairwise matching strategy that PointContrast adopts allows one frame to be matched multiple times. As a result, each frame can be encoded multiple times in one step, adding to redundancy in training.
    \vspace{-2mm}
    \item \textit{Low learning efficiency.} In one training step, the frame matching strategy only allows the framework to process several views of a single scene. Therefore the amount of information that PointContrast can process in one step is rather limited, and the overall time for one training cycle is also notably high.
    \vspace{-2mm}
    \item \textit{Dependency on raw RGB-D frames.} The whole framework is built on the assumption that RGB-D videos are available, yet this is not true for every publicly-available point cloud dataset. Even when available, the storage cost of RGBD frames is also significantly higher than the reconstructed point cloud data.
    \vspace{-2mm}
\end{itemize}
Consequently, pre-training frameworks~\cite{xie2020pointcontrast,hou2021exploring} based on matching frames as contrastive samples require enormous computing and storage resources. For example, PointContrast sub-samples RGB-D scans from the raw ScanNet videos every 25 frames, consuming $\sim$30 times the storage of the processed point clouds, and takes 80 GPU hours to process an epoch of 1500 scenes. Even at such cost, our experiments in \tabref{tab:ablation_view_generation} verify that the raw RGB-D frames cannot bring additional information over the processed point clouds to achieve a better representation.

\textit{We will explore the possibility of pre-training on the point clouds directly.}

\begin{figure*}[!t]\centering
\includegraphics[width=0.95\linewidth]{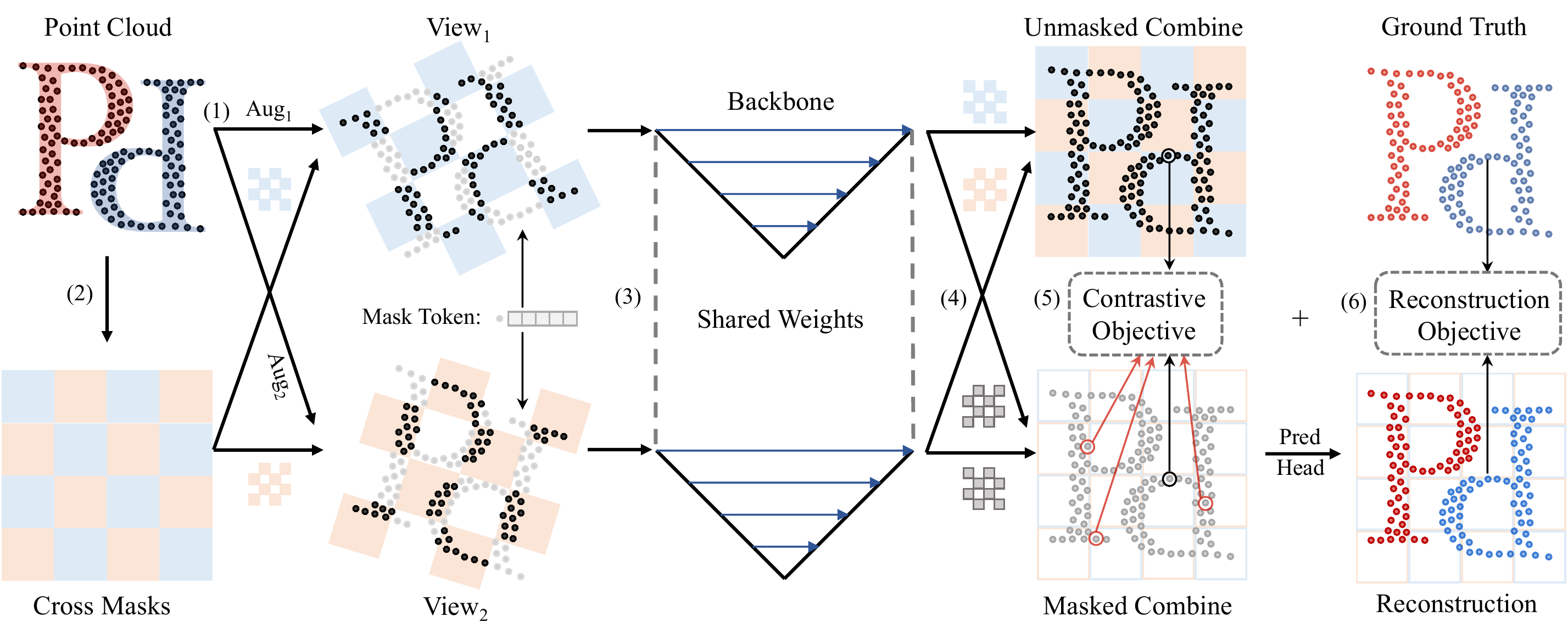}
\caption{\textbf{Our MSC framework}. (1) Generating a pair of contrastive views with a well-curated data augmentation pipeline consisting of photometric, spatial, and sampling augmentations. (2) Generating a pair of complementary masks and applying them to the pair of contrastive views. Replacing masked point features with a learnable mask token vector. (3) Extracting point representation with a given U-Net style backbone for point cloud understanding. (4) Reassembling masked contrastive views to masked points combination and unmasked points combination. (5) Matching points share similar positional relationships in the two views as positive sample pairs and computing InfoNCE loss to optimize contrastive objective. (6) Predicting masked point color and normal and computing Mean Squared Error loss and Cosine Similarity Loss with ground truth respectively, to optimize the reconstruction objective.}
\label{fig:framework}
\end{figure*}

\mypara{What's the revelation behind mode collapse?} \\
Mode collapse, defined as the phenomenon that all features collapse to a single vector, remains an unsolved problem accompanying the development of 3D representation learning~\cite{choy2019fully,xie2020pointcontrast}.
To alleviate this problem, PointContrast introduced InfoNCE loss~\cite{oord2018representation}, which has been shown to stabilize training, to replace the hardest-contrastive loss. Yet, the problem of mode collapse can still occur when the amount of training data and the length of the training schedule increase.
Given the empirical conclusion of 2D contrastive learning~\cite{chen2021exploring}, the occurrence of mode collapse is unusual under the premise that a large number of negative samples are already adopted. 
Interestingly, we notice that the mean negative pair cosine similarity  of previous works is mostly close to 0, indicating that the negative samples are mostly easy and thus have little penalty towards the trivial solution. Although the InfoNCE loss alleviates this problem with an alternated optimization objective, we argue that a more desirable solution can be achieved by raising the difficulty of the unsupervised pretext task.

\textit{We will further raise the difficulty of the pretext task to solve the mode collapse problem.}

\section{Approach}
Based on the analysis of the stumbling blocks for large-scale pre-training in \secref{sec:pilot}, we first introduce our optimized contrastive learning design in \secref{sec:cl} to make the process more efficient. Then we solve the long-term problem of mode collapse with an additional reconstructive learning design in \secref{sec:rl}. The final optimization target is described in \secref{sec:loss}. Combining these exquisite designs, we build the whole framework, namely \textit{Masked Scene Contrast} (MSC), and a visual illustration of our MSC is available in \figref{fig:framework}.
\subsection{Contrastive Learning}
\label{sec:cl}
\mypara{Framework.}
Different from the previous protocol of matching RGB-D frames decomposed from indoor scenes, our contrastive learning framework directly operates on the point cloud data. Given a point cloud $\mX = (\mP, \mC)$, where $\mP \in \mathbb{R}^{n \times 3}$ represents the spatial features (coordinate) of the points and $\mC \in \mathbb{R}^{n \times 3}$ represents the photometric features (color) of the points, the contrastive learning framework can be summarized as follows:

\begin{itemize}
    \vspace{-2mm}   
    \item \textit{View generation.} For a given point cloud $\mX$, we generate query view $\mX_r$ and key view $\mX_k$ of the original point cloud with a sequence of stochastic data augmentations, which includes photometric, spatial, and sampling augmentations.
    \vspace{-2mm}
    \item \textit{Feature extraction.} Encoding point cloud features $\mF_r$ and $\mF_k$ with a U-Net style backbone $\zeta(\cdot)$ to $\hat{\mF}_r$ and $\hat{\mF}_k$ respectively.
    \vspace{-2mm}
    \item \textit{Point matching.} The positive samples of contrastive learning are point pairs with close spatial positions in the two views. For each point belonging to the query view, we calculate the correspondence mapping $\sP = \{(i, j)\}_{n'}$ to points of the key view. If $(i, j) \in \sP$ then point $(\vp_i, \vc_i)$ and point $(\vp_j, \vc_j)$ constructs a pair across two views.
    \vspace{-2mm}
    \item \textit{Loss computation.} Computing the contrastive learning loss on the representation of two views $\hat{\mF}_r$ and $\hat{\mF}_k$ and the correspondence mapping $\sP$. An encoded query view should be similar to its key view.
    \vspace{-2mm}
\end{itemize}

\mypara{Data augmentation.} As a pioneering work in image contrastive learning, SimCLR~\cite{chen2020simple} reveals that a well-curated data augmentation pipeline is crucial for learning strong representations. Unlike supervised learning, contrastive learning requires much stronger data augmentations to prevent trivial solutions.
However, an effective data augmentation recipe is still absent in 3D representation learning. The frame matching scheme in prior works~\cite{xie2020pointcontrast,hou2021exploring} simply applies a randomly rotating operator to contrastive targets. Even if the RGB-D frames can be viewed as a natural random crop, the augmentation space is still far from diverse enough. The pretext task it forms is not yet challenging enough to facilitate the contrastive learning framework to learn robust representations for downstream tasks.

As presented in \figref{fig:scene_augmentation}, our well-designed stochastic data augmentation pipeline includes photometric augmentations, spatial augmentations, and sampling augmentations.
Inspired by the advanced photometric augmentation validated by our 2D counterparts~\cite{chen2020simple,grill2020bootstrap}, we further strengthen the photometric augmentation component introduced by Choy \etal~\cite{choy20194d} with random brightness, contrast, saturation, hue, and gaussian noise jittering for photometric augmentation. Besides that, random rotating, flipping, and scaling constitute our spatial augmentations, and the sampling augmentation is composed of random cropping and grid sampling.

Empirically, the order of data augmentations is also a key component of our recipe.
For example, grid sampling after random rotation leads to cross grids for sampling, which further increases the distinction between contrastive views and has a better augmentation effect. The specific data augmentation settings are available in the appendix, and a comparison with previous methods is presented in \figref{fig:view_generation}.

\begin{figure}[t]\centering
\includegraphics[width=\linewidth]{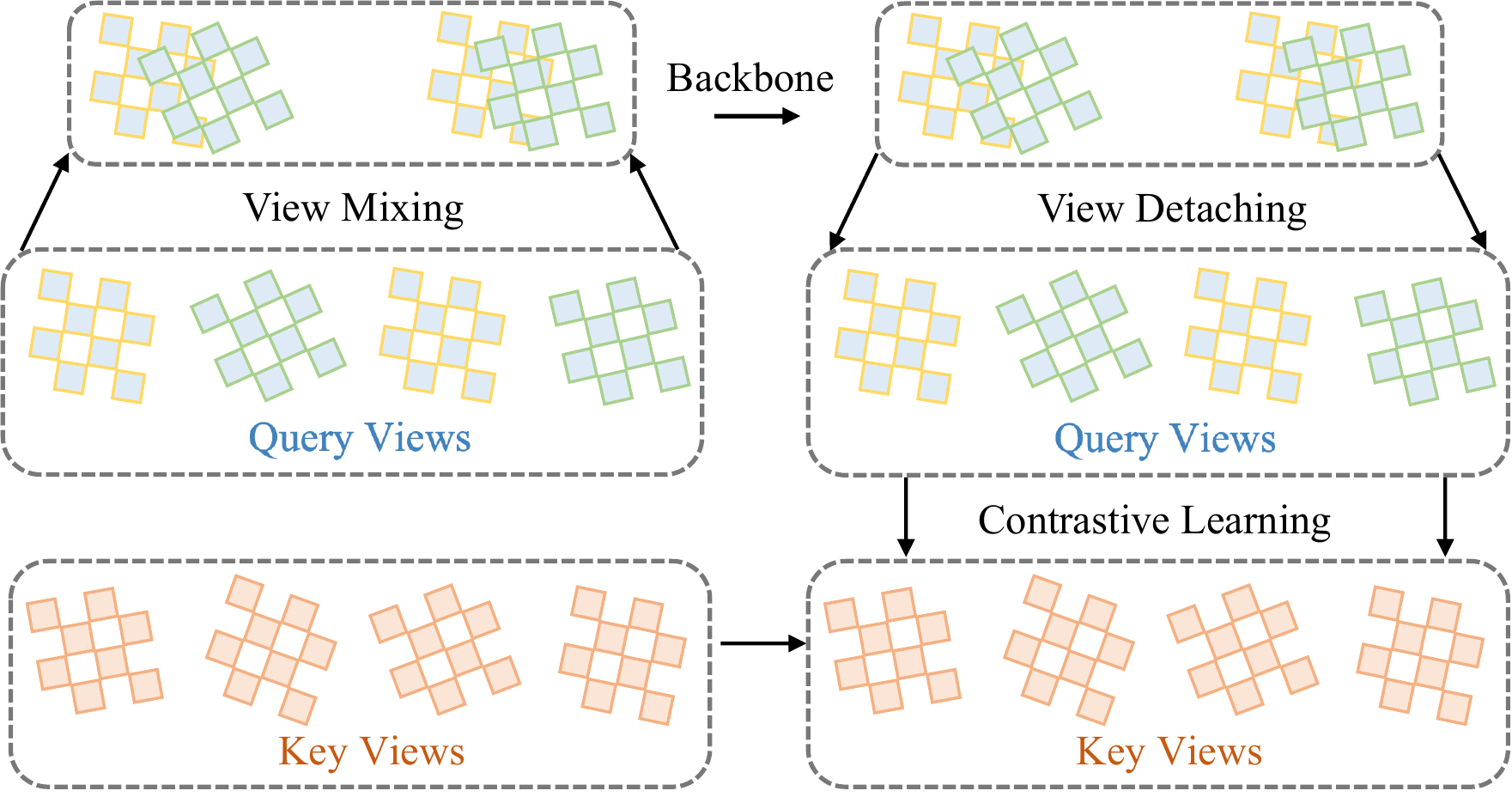}
\caption{\textbf{View Mixing}. Randomly mix up query views while keeping key views unmixed for a given batch of pairwise contrastive views. Detaching mixed query view after feature extraction for contrastive comparison with matched key views.}
\label{fig:mixing}
\end{figure}

\mypara{View mixing.} Recently, Nekrasov \etal~\cite{Nekrasov21mix3d} proposes a data augmentation technique for 3D understanding models by mixing two scenes as a hybrid training sample, which can significantly suppress model overfitting. Inspired by the mixing mechanism, we integrate the logic of mixing as part of the contrastive learning objective. As illustrated in \figref{fig:mixing}, for a batch of pairwise views, we randomly mix up the query views while maintaining the key views unchanged before the \textit{feature extraction} process. The simple operation can effectively increase the robustness of the backbone and improve the robustness of the point cloud representation.

\mypara{Contrastive target.} We follow the design of PointContrast on the contrast target and apply InfoNCE loss to the matched points. Given correspondence mapping $\sP = \{(i, j)\}_{n'}$  produced by \textit{point matching} and points representation $\hat{\mF}_r$ and $\hat{\mF}_k$ embedded during \textit{feature extraction}, the contrastive loss is:
\begin{align}
    s_{ij} &= \frac{\vf_{i'}^{rT} \vf_{j'}^{k}}{||\vf_{i'}^{rT}||\cdot||\vf_{j'}^{k}||},
    \label{eq:cosine-similarity}
    \\
    \mathcal{L}_{\text{InfoNCE}} &= \sum_i^n - \text{log} \frac{\text{exp}(s_{ii} / \tau)}{\sum_j^n \text{exp}(s_{ij} / \tau)},
    \label{eq:info-nce-loss}
\end{align}
note that $\mS=\{s_{ij}\} \in \mathbb{R}^{n\times n}$ is the pairwise cosine similarity matrix between positive samples and negative samples, while $\tau$ is the temperature factor scaling cosine similarity. In practice, we control temperature factor $\tau$ as 0.4, which is the same as previous works~\cite{xie2020pointcontrast,hou2021exploring}.

\subsection{Reconstructive Learning}
\label{sec:rl}
As is mentioned in \secref{sec:pilot}, one of the stumbling blocks for large-scale representation is mode collapse, and our solution is to scale up the difficulty of the unsupervised pre-training task. Motivated by the success of masked image modeling~\cite{he2022masked, xie2022simmim} in 2D representations, we propose masked point modeling, which can be naturally integrated into our contrastive learning framework. Benefiting from this design, our framework can fully use non-overlapped regions of contrastive views that cannot be utilized by contrastive learning.

\mypara{Contrastive cross mask.}
The key design that enables additional construction learning in our contrastive learning framework is the contrastive cross mask. For a given query view and key view of a single point cloud, we partite the unioned point set into non-overlapping grid partitions by their original position before spatial augmentation. Given a mask rate $r$ range from 0 to 0.5, we randomly generate a pair of masks $\mM_r, \mM_k \in \mathbb{R}^{1\times n_{r,k}}$, in which there are no shared masked patches. Then, we follow the practice of SimMIM~\cite{xie2022simmim} to apply the pair of masks to the two views respectively by replacing the input feature with a learnable mask token vector $\vt \in \mathbb{R}^c$. Consequently, the feature extraction process can be rewritten as follows:
\begin{align}
    \hat{\mF}_{r,k} &= \zeta((1 - \mM_{r,k}) \mF_{r,k} + \mM_{r,k}\mT_{r,k}),
    \label{eq:feature-extract-with-mask}
\end{align}
where $\mT_{r,k}\in\mathbb{R}^{n_r,n_k\times c}$ is the expand matrix of mask token vector $\vt$ to fit the feature dimensions.

\mypara{Reconstruction target.}
The features of the point cloud are composed of two parts, the coordinates that determine the geometric structure and the colors that represent the texture features. We build up reconstruction targets for the two groups of features separately.

The reconstruction of point cloud texture is straightforward, we predict the photometric value of each point with a linear projection. We compute the mean squared error (MSE) between the reconstructed and original color of masked points as the color reconstruction loss:
\begin{align}
    \mathcal{L}_{\text{c}} = \frac{\sum_i^{n_r}m_i^r||\vx_i^r - \hat{\vx}_i^r||_2^2 + \sum_i^{n_k}m_i^k||\vx_i^k - \hat{\vx}_i^k||_2^2}{n_r' + n_k'},
    \label{eq:color-reconstruction-loss}
\end{align}
where $n_r'$ and $n_k'$ represent the number of mask points belonging to refer view and key view, $m_r^i$ and $m_k^i$ mean the $i$-th element of $M_r$ and $M_k$ respectively.

Point coordinates play an important role in describing the geometric structure of point clouds, and it is worth noting that directly reconstructing the coordinates of masked points is not reasonable since masked points are only sampled from 3D object surface rather than the continuous surface itself. Reconstructing points coordination would lead to an overfitted representation. To overcome the challenge, we introduce the concept of surfel reconstruction. Surfel is an abbreviation for a \textit{surface element} or \textit{surface voxel} in the discrete topology literature~\cite{herman1992discrete} and primitives rendering~\cite{pfister2000surfels}. For each masked point, we reconstruct the normal vector of the corresponding surfel and compute the  mean cosine similarity between estimations and surfel normals as a contrastive loss:
\begin{align}
    \mathcal{L}_{\text{n}} = \frac{\sum_i^{n_r}m_i^r\vx_i^{rT}\hat{\vx}_i^{r} + \sum_i^{n_k}m_i^k\vx_i^{kT}\hat{\vx}_i^{k}}{n_r' + n_k'},
    \label{eq:normal-reconstruction-loss}
\end{align}
where $n_r'$ and $n_k'$ represent the number of mask points belonging to refer view and key view, $m_r^i$ and $m_k^i$ mean the $i$-th element of $M_r$ and $M_k$ respectively.

\subsection{Loss Function}
\label{sec:loss}
Our framework combines the contrastive target, the color reconstruction target, and the surfel reconstruction to make the unsupervised task more scalable. The overall loss function is a weighted sum of \eqref{eq:info-nce-loss}, \eqref{eq:color-reconstruction-loss}, and \eqref{eq:normal-reconstruction-loss} which is written as follow:
\begin{align}
    \mathcal{L}_{\text{overall}} = \mathcal{L}_{\text{InfoNCE}} + \lambda_c\mathcal{L}_{\text{c}} + \lambda_n\mathcal{L}_{\text{n}},
    \label{eq:combined-loss}
\end{align}
where $\lambda_c$ and $\lambda_n$ are the weight parameters that balance the three loss components. Empirically we find that performance is robust to the choice of weight parameters, and we make $\lambda_c = \lambda_n = 1$ in practice.

\begin{figure}[t]\centering
\includegraphics[width=.98\linewidth]{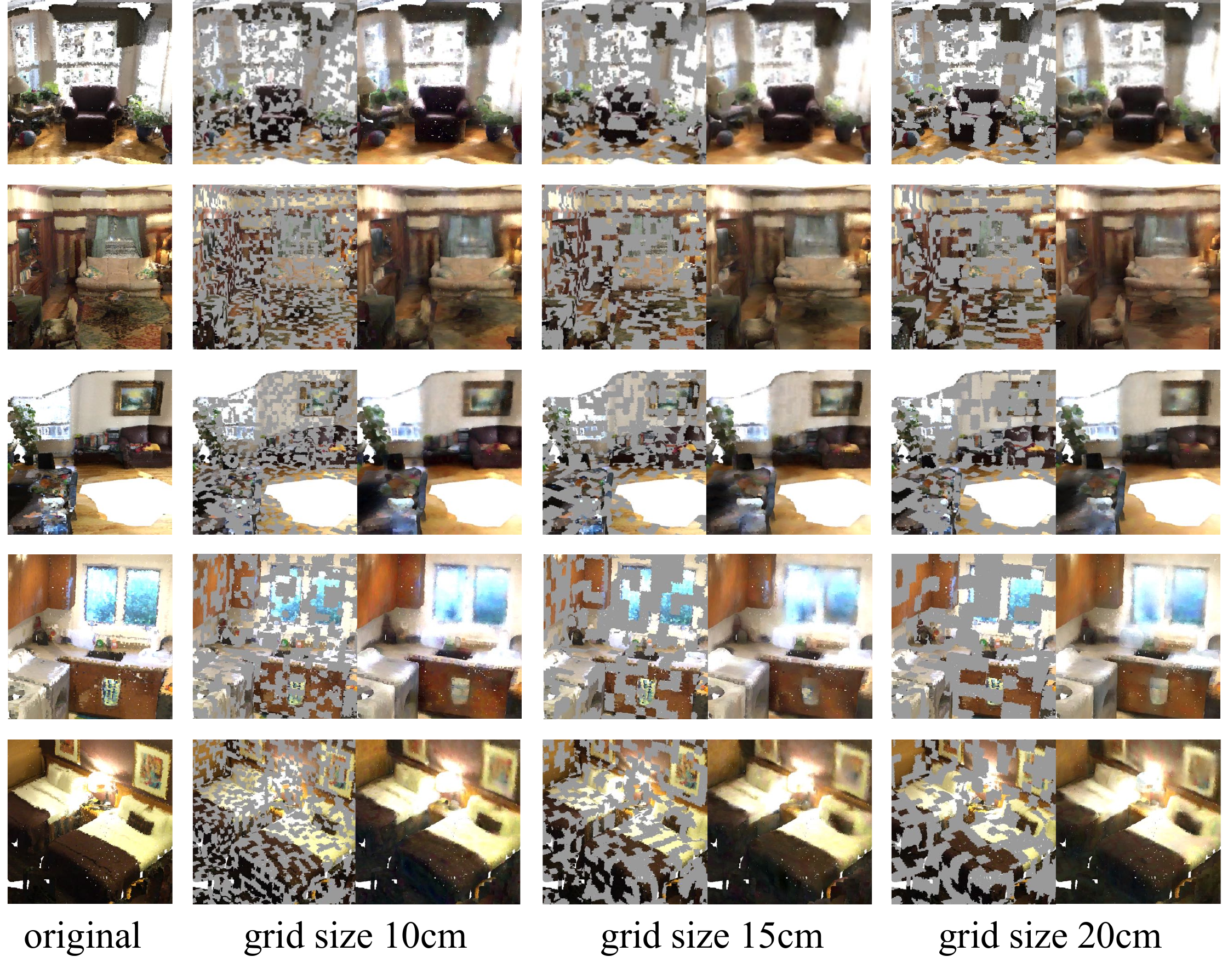} \\
\caption{\textbf{Masked scenes and color reconstructions}. We visualize one of the cross masks of each scene with a mask rate of 50\% (left) and color reconstruction of the masked point combinations (right). We pre-train our MSC with a mask patch size of 10cm and generalize to results to different mask sizes. Compared with the original point clouds, the loss of detail cannot be avoided, while boundary and texture are well preserved by our model.}
\label{fig:vis_mask}
\end{figure} 

\section{Experiments}
\begin{table*}[t!]
    \subfloat[
        \textbf{View generation.}Views produced by our enhanced data augmentation are stronger than the original RGB-D frames. Scene-level views can significantly speed up pre-training and make contrastive learning more effective. The performance can be further boosted with additional masked point modeling.
        \vspace{1mm}
        \label{tab:ablation_view_generation}
    ]{
        \begin{minipage}{0.98\linewidth}{\begin{center}
            \tablestyle{6pt}{1.02}
            \begin{tabular}{lccccccccc}
\toprule
View generation methods &Pre-train data &Storage &Batch size &Iters &Epochs &FT mIoU (\%) &Hours (h) &Speedup \\
\midrule
Frame matching (PointContrast~\cite{xie2020pointcontrast}) &ScanNet Raw &500G &32 &100k &5 &74.0 &48 & 1.0$\times$ \\
Scene augmentation w/o mask (ours) &ScanNet v2 &20G &32 &30k &600 &74.4 &\textbf{11} &\textbf{4.4$\times$} \\
\cellcolor[HTML]{efefef}Scene augmentation w mask (ours) &ScanNet v2 &20G &32 &30k &600 &\textbf{75.0} &14 &3.4$\times$ \\
\bottomrule
\end{tabular}

        \end{center}}\end{minipage}
    } \\
    \centering
    \subfloat[
        \textbf{Number of positive pairs.} A larger amount of sampled positive pairs are necessary for scene-level views.
        \label{tab:ablation_ft_bs}
    ]{
        \begin{minipage}{0.29\linewidth}{\begin{center}
            \tablestyle{4pt}{1.02}
            \begin{tabular}{cccc}\toprule
\#Pos pairs &PC~\cite{xie2020pointcontrast} &MSC~(ours) \\\midrule
1024 &73.8 &74.3 \\
2048 &74.0 &74.5 \\
4098 &73.7 &74.9 \\
\cellcolor[HTML]{efefef}8192 &73.9 &\textbf{75.0} \\
\bottomrule
\end{tabular}
        \end{center}}\end{minipage}
    }
    \hspace{5mm}
    \subfloat[
        \textbf{Data augmentation.} The combination of spatial and photometric augmentation makes the view generation pipeline comes to work.
        \label{tab:ablation_augmentation}
    ]{
        \begin{minipage}{0.29\linewidth}{\begin{center}
            \tablestyle{4pt}{1.02}
            \begin{tabular}{cccc}\toprule
Spatial &Photometric &FT mIoU (\%) \\\midrule
w/o aug &w/o aug &72.1 \\
w aug &w/o aug &73.4 \\
w/o aug &w aug &72.8 \\
\cellcolor[HTML]{efefef}w aug &\cellcolor[HTML]{efefef}w aug &\textbf{74.4} \\
\bottomrule
\end{tabular}
        \end{center}}\end{minipage}
    }
    \hspace{5mm}
    \subfloat[
        \textbf{View mixing.} Randomly mixing query views while leaving key views unmixed is a sweet point.
        \label{tab:ablation_mix}
    ]{
        \begin{minipage}{0.29\linewidth}{\begin{center}
            \tablestyle{4pt}{1.02}
            \begin{tabular}{cccc}\toprule
Query view &Key view &FT mIoU (\%) \\\midrule
w/o mix &w/o mix &74.1 \\
\cellcolor[HTML]{efefef}w mix &\cellcolor[HTML]{efefef}w/o mix &\textbf{74.4} \\
w/o mix &w mix &74.2 \\
w mix &w mix &73.7 \\
\bottomrule
\end{tabular}
        \end{center}}\end{minipage}
    } \\
    \centering
    \vspace{1.5mm}
    \subfloat[
        \textbf{Cross mask.} Masks containing shared masked patches have a negative influence on contrastive learning.
        \label{tab:ablation_cross_mask}
    ]{
        \begin{minipage}{0.29\linewidth}{\begin{center}
            \tablestyle{4pt}{1.02}
            \begin{tabular}{cccc}\toprule
Mask &Task &FT mIoU (\%) \\\midrule
w/o cross &w/o contrast &74.1 \\
w cross &w/o contrast &74.4 \\
w/o cross &w contrast &74.7 \\
\cellcolor[HTML]{efefef}w cross &\cellcolor[HTML]{efefef}w contrast &\textbf{75.0} \\
\bottomrule
\end{tabular}
        \end{center}}\end{minipage}
    }
    \hspace{5mm}
    \subfloat[
        \textbf{Mask grid size.} Our design works with mask patches with a grid size larger than 0.1m, and we consider 0.15m as a default setting.
        \label{tab:ablation_mask_grid_size}
    ]{
        \begin{minipage}{0.29\linewidth}{\begin{center}
            \tablestyle{4pt}{1.02}
            \begin{tabular}{ccc}\toprule
Mask grid size (m) &FT mIoU (\%) \\\midrule
0.05 &74.3 \\
0.1 &75.0 \\
\cellcolor[HTML]{efefef}0.15 &\textbf{75.0} \\
0.2 &74.8 \\
\bottomrule
\end{tabular}
        \end{center}}\end{minipage}
    }
    \hspace{5mm}
    \subfloat[
        \textbf{Reconstruction target.} Both targets have a positive effect, while color reconstruction has a dominant impact on indoor scenes.
        \label{tab:ablation_reconstruction_target}
    ]{
        \begin{minipage}{0.29\linewidth}{\begin{center}
            \tablestyle{4pt}{1.02}
            \begin{tabular}{cccc}\toprule
Color &Normal &FT mIoU (\%) \\\midrule
w/o &w/o &74.4 \\
w &w/o &74.9 \\
w/o &w &74.6 \\
\cellcolor[HTML]{efefef}w &\cellcolor[HTML]{efefef}w &\textbf{75.0} \\
\bottomrule
\end{tabular}
        \end{center}}\end{minipage}
    }
    \vspace{-2mm}
    \caption{\textbf{Ablation experiments.} We adopt \textit{SparseUNet} and \textit{efficient} pre-training on ScanNet~\cite{dai2017scannet} point cloud data to ablate our designs. We report fine-turning (FT) mIoU (\%) results on ScanNet 20 classes semantic segmentation as the default metric. If not specified, the default setting is as follows: the pre-training period is 600 epochs, the masking ratio is 30\% and the masked patch has a grid size of 0.15m in the real-world space, view mixing probability is 0.8. All of our designs are enabled by default. Default settings are marked in \colorbox{gray}{gray}.} 
    \label{tab:ablation}
    \vspace{-3mm}
\end{table*}

We conduct extensive experimental evaluations to validate the capability of our framework, built upon the point cloud perception codebase \textit{Pointcept}~\cite{pointcept2023}. We first ablate our designs with an efficient pre-training pipeline that only utilizes ScanNet point cloud in \secref{sec:ablation}, while without compromising performance. Then we explore large-scale pre-training across multiple datasets and compare our performance with previous results in \secref{sec:results_comparison}.

\subsection{Main Properties}
\label{sec:ablation}
We ablate the main designs and intriguing properties of our MSC in \tabref{tab:ablation}, and the default setting is available in the caption. We enable \textit{efficient pre-training} by introducing our view generation pipeline, which is ablated in \tabref{tab:ablation_view_generation}. All of our ablation experiments only require 20G ScanNet point cloud data and around 14 hours pertaining on a single machine containing 8 NVIDIA RTX3090.

\mypara{View generation.}
In \tabref{tab:ablation_view_generation}, we show the results with different generation strategies of contrastive views. Unlike PointContrast~\cite{xie2020pointcontrast}, which builds on the raw RGB-D frames, our strategy directly utilizes the scene-level point cloud.
This significantly reduces storage requirements (96\% less) and allows more efficient use of the training data.
With a 30\% equivalent number of training iterations, our method can attain 120$\times$ training epochs, making full use of the training data for the first time.
This results in 0.4 points higher FT mIoU with 4.4$\times$ speedup.
When combined with an additional mask point modeling strategy, the performance can be further boosted by 0.6 points, yet still with a notable speedup.

\mypara{Number of positive pairs.}
In \tabref{tab:ablation_ft_bs}, we show the effects of different numbers of positive pairs. Our method sees consistent improvements with an increasing number of positive pairs, while for PointContrast~\cite{xie2020pointcontrast}, the information in a large batch of positive pairs cannot be effectively utilized. 
Our intuition is that the ability to process scene-level views, which are larger in scale and contain much more information than frame-level views, enables our method to take advantage of a larger number of positive pairs.

\mypara{Data augmentation.}
In \tabref{tab:ablation_augmentation}, we analyze the effect of different data augmentation combinations.
Adopting either spatial augmentation or photometric augmentation leads to sub-optimal performance, and the combination of both helps make our view generation pipeline come to work.
Concerning relative significance, since the spatial augmentation is highly coupled with the masking strategy, the gain from it is higher than the photometric augmentation.
But still, both are necessary.

\mypara{View mixing.}
In \tabref{tab:ablation_mix}, we explore different configs for the view mixing strategy.
Randomly mixing the query views while leaving the key views unchanged yields the best performance.
Our intuition is that the key views, which form the vocabulary of the InfoNCE loss, should be relatively stable, thus reducing the ambiguity of the learning target.
In contrast, introducing more diversity to the queries can be more helpful.

\begin{figure}[t]\centering
\includegraphics[width=.95\linewidth]{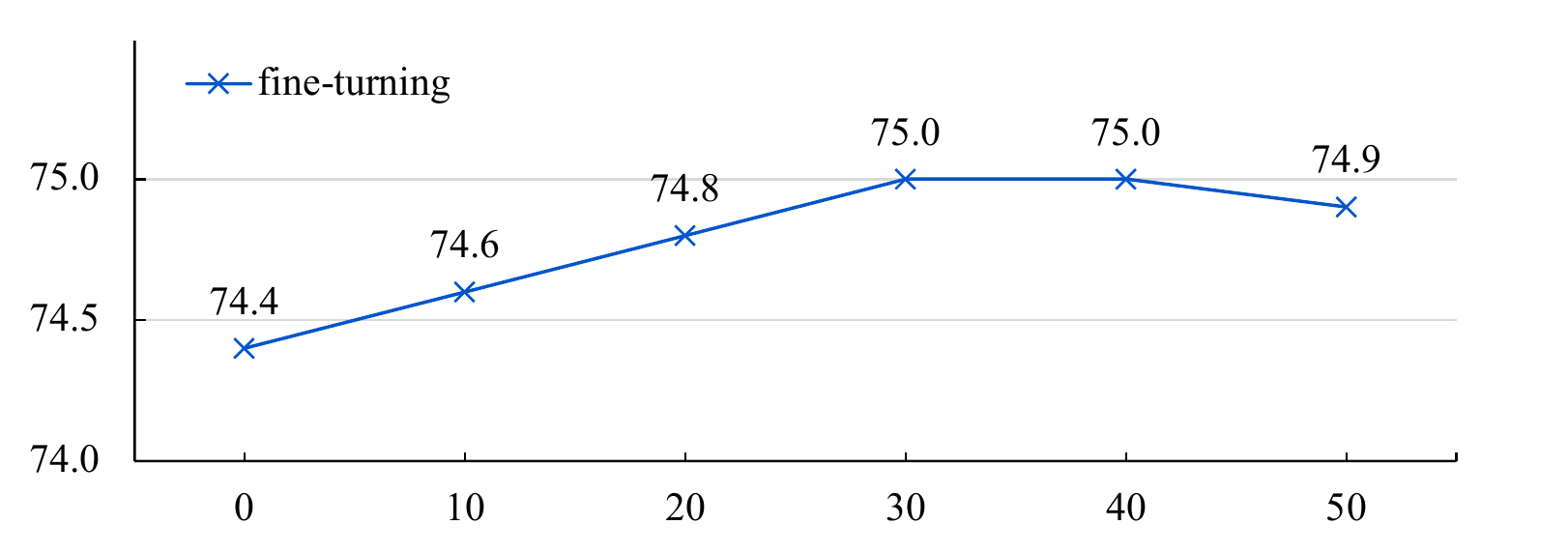} \\
\scriptsize masking ratio (\%) \\
\caption{\textbf{Masking ratio}. A masking ratio ranging from 30\% to 40\% works well with our design, and a higher mask rate negatively influences contrastive learning. The y-axes represent ScanNet semantic segmentation validation mIoU (\%).}
\vspace{-6mm}
\label{fig:ablation_mask_ratio}
\end{figure}

\mypara{Cross mask.}
In \tabref{tab:ablation_cross_mask}, we study the cross mask strategy.
This strategy ensures that the two augmented views have no overlapping tokens.
As reported in the table, whether with or without the contrastive learning target, this strategy ensures fewer shortcuts to the task and enables consistent downstream improvements.

\mypara{Mask grid size.}
In \tabref{tab:ablation_mask_grid_size}, we show the results ablating the grid size for producing masks on point clouds.
Our design works with a grid size larger than 0.1m, and we consider 0.15m as a default setting. As shown in \figref{fig:vis_mask}, our reconstruction module is robust to extending mask grid size, which indicates that high quilty representation is captured during pre-training.

\mypara{Masking ratio.}
In \figref{fig:ablation_mask_ratio}, we depict the effect of the ratio of masked tokens.
A masking ratio ranging from 30\% to 40\% works well with our design, and a higher masking rate has a negative impact on the overall performance.
This varies from the conclusion of MAE~\cite{he2022masked}, in which a higher masking ratio of 75\% achieves top performance.
Our hypothesis is that the contrastive learning objective built on the masked point clouds might favor a lower masking ratio, and the results in \figref{fig:ablation_mask_ratio} reflect a trade-off between the contrastive learning objective and the reconstructive objective.
And pure reconstructive learning on point clouds is an interesting direction for future explorations.

\mypara{Reconstruction target.}
In \tabref{tab:ablation_reconstruction_target}, we ablate the effects of two components of our reconstruction target: color reconstruction and normal reconstruction. 
Given the premise that indoor scenes are used, both targets show a positive effect on overall performance, while color reconstruction has a higher impact.
The intuition is that the difference in texture reflected by color has a higher impact on the task of semantic segmentation, while the normal is helpful but has less influence (consider the same task on a 2D image).

\begin{table}[t!]
    \centering
    \subfloat[
        \textbf{Semantic segmentation.} We conduct pre-training on SparseUNet and compare semantic segmentation mIoU~(\%) results on ScanNet and ScanNet200~\cite{rozenberszki2022language} validation set.
        \vspace{1mm}
        \label{tab:results_sem_seg}
    ]{
        \begin{minipage}{0.95\linewidth}{\begin{center}
            \tablestyle{2.5pt}{1.02}
            \begin{tabular}{cc|ccccc}\toprule
\multirow{2}{*}{Datasets} &\multirow{2}{*}{Backbones} &\multicolumn{4}{c}{Semantic Seg. (mIoU)} \\\cmidrule{3-6}
& &SC &PC~\cite{xie2020pointcontrast} &CSC~\cite{hou2021exploring} &\cellcolor[HTML]{efefef}MSC~(ours) \\\midrule
ScanNet &SparseUNet &72.2 &74.1~\tiny{\textcolor{darkgray}{(+1.9)}} &73.8~\tiny{\textcolor{darkgray}{(+1.6)}} &\textbf{75.5}~\tiny{\textcolor{darkgreen}{(+3.3)}} \\
ScanNet200 &SparseUNet &25.0 &26.2~\tiny{\textcolor{darkgray}{(+1.2)}} &26.4~\tiny{\textcolor{darkgray}{(+1.4)}}&\textbf{28.8}~\tiny{\textcolor{darkgreen}{(+3.8)}} \\
\bottomrule
\end{tabular}

        \end{center}}\end{minipage}
    }
    \\
    \centering
    \subfloat[
        \textbf{Instance segmentation.} We conduct pre-training on SparseUNet and compare instance segmentation mAP@0.5~(\%) results driven by \textit{PointGroup}~\cite{jiang2020pointgroup} on ScanNet and ScanNet200~\cite{rozenberszki2022language} validation set.
        \label{tab:results_ins_seg}
    ]{
        \begin{minipage}{0.95\linewidth}{\begin{center}
            \tablestyle{2.5pt}{1.02}
            \begin{tabular}{cc|ccccc}\toprule
\multirow{2}{*}{Datasets} &\multirow{2}{*}{Backbones} &\multicolumn{4}{c}{Instance Seg. (mAP@0.5)} \\\cmidrule{3-6}
& &SC &PC~\cite{xie2020pointcontrast} &CSC~\cite{hou2021exploring} &\cellcolor[HTML]{efefef}MSC~(ours) \\\midrule
ScanNet &SparseUNet &56.9 &58.0~\tiny{\textcolor{darkgray}{(+1.1)}} &59.4~\tiny{\textcolor{darkgray}{(+2.5)}} &\textbf{59.6}~\tiny{\textcolor{darkgreen}{(+2.7)}}\\
ScanNet200 &SparseUNet &24.5 &24.9~\tiny{\textcolor{darkgray}{(+0.4)}} &25.2~\tiny{\textcolor{darkgray}{(+0.7)}} &\textbf{26.8}~\tiny{\textcolor{darkgreen}{(+2.3)}}\\
\bottomrule
\end{tabular}

        \end{center}}\end{minipage}
    }
    \\
    \vspace{1mm}
    \subfloat[
        \textbf{Data efficiency.} We follow the ScanNet Data Efficient benchmark and compare the validation results SparseUNet with previous methods.
        \label{tab:results_data_efficient}
    ]{
        \begin{minipage}{0.95\linewidth}{\begin{center}
            \tablestyle{2.4pt}{1.02}
            \begin{tabular}{c|ccccc|c|ccccc}\toprule
LR &\multicolumn{4}{c}{Semantic Seg.} & &LA &\multicolumn{4}{c}{Semantic Seg.} \\\cmidrule{1-5}\cmidrule{6-11}
Pct. &SC &CSC &VIBUS &\cellcolor[HTML]{efefef}MSC & &Pts. &SC &CSC &VIBUS &\cellcolor[HTML]{efefef}MSC \\\midrule
100\% &72.2 &73.8 &- &\textbf{75.3} & &Full &72.2 &73.8 &- & \textbf{75.3} \\
1\% &26.0 &28.9 &28.6 &\textbf{29.2} & &20 &41.9 &55.5 &61.0 & \textbf{61.2}\\
5\% &47.8 &49.8 &47.4 &\textbf{50.7} & &50 &53.9 &60.5 &65.6 & \textbf{66.8}\\
10\% &56.7 &59.4 &60.5 &\textbf{61.0} & &100 &62.2 &65.9 &68.9 & \textbf{69.7}\\
20\% &62.9 &64.6 &64.8 &\textbf{64.9} & &200 &65.5 &68.2 &69.6 & \textbf{70.7}\\
\bottomrule
\end{tabular}
        \end{center}}\end{minipage}
    }
    \vspace{-2mm}
    \caption{\textbf{Results comparison.} We adopt cross-dataset pre-training utilizing ScanNet and ArkitScenes point cloud scenes for comparison of downstream task results. The pre-training setting is fixed as the default described in \tabref{tab:ablation}. More specific pre-training details are available in the Appendix. \textit{SC} denotes train from scratch.}
    \vspace{-5mm}
    \label{tab:results}
\end{table}

\subsection{Results Comparison}
\label{sec:results_comparison}
In this section, we extend the scale of pre-training by merging multiple datasets and compare downstream task fine-turning performance with previous unsupervised pre-training frameworks~\cite{xie2020pointcontrast, hou2021exploring}.
Specifically, we adopt the default model setting ablated in \secref{sec:ablation} and train on both ScanNet~\cite{dai2017scannet} and ArkitScenes~\cite{dehghan2021arkitscenes} point clouds, extending pre-training assets from 1,513 scenes to 6,560 scenes.

\mypara{Semantic segmentation.}
In \tabref{tab:results_sem_seg}, we report the semantic segmentation results on ScanNet and ScanNet200~\cite{rozenberszki2022language} benchmark with SparseUNet and compare them with previous results.
Our improvements are consistent and significant with larger pre-training assets for both benchmarks.
With SparseUNet backbone, we outperform the current state-of-art pretraining framework by 1.7 points on ScanNet and 2.4 points on ScanNet200. Meanwhile, it is worth noting that driven by powerful MSC, we set a new best validation result on ScanNet semantic segmentation and pushed the previous SOTA to 75.5\% with a baseline model.

\mypara{Instance segmentation.}
In \tabref{tab:results_ins_seg}, we report the instance segmentation results on ScanNet and ScanNet200~\cite{rozenberszki2022language} with SparseUNet backbone driven by \textit{PointGroup}~\cite{jiang2020pointgroup}. Comparing them with previous results, we still see consistent improvements. Specifically, our framework achieves 59.6\% on the ScanNet validation set, which is 3.3 points higher than training from scratch and 0.2 points promotion compared with the previous state-of-art performance. The boost is more significant on ScanNet200, which is 1.6 points higher than the previous SOTA.

\mypara{Data efficiency.} In \tabref{tab:results_data_efficient}, we compare the ScanNet Data Efficient~\cite{hou2021exploring} results with previous methods. Our MSC shows consistently superior performance even compared with the latest data-efficient learning framework\cite{tian2022vibus}.

\section{Conclusion and Discussion}
In this paper, we tackle the problem of scalable unsupervised 3D representation learning. To this end, we present Masked Scene Contrast (MSC), an efficient, effective, and scalable framework that directly operates on scene-level views with contrastive learning and masked point modeling. Benefiting from the efficient scene-level point cloud processing pipeline and the effective training objectives, our method harvests high efficiency and superior generality, and enables large-scale pre-training across multiple datasets.

The key factor that empowers our method's scalability to larger-scale pre-training lies in the efficient pipeline that can directly learn from point cloud data, rather than the raw RGB-D frames. The efficiency, however, does not only mean processing data at scale. When only the standard dataset ScanNet is used for pre-training, our method still achieves uncompromised performance, yet with at least 3$\times$ speedup over previous works. This is especially meaningful considering the exhaustively long experimental time in the pre-training community, and it can better facilitate the verification of new ideas in future works.

It should also \textbf{}be noted that given the limit in computing resources, and the absence of a pre-training dataset that is sustainably at scale, the scalability of our method is not fully presented. In other words, our method opens the possibility of large-scale pre-training on 3D point cloud data for the first time, and we call for a large-enough 3D scene dataset that can fully unleash this potential. We hope our method can inspire future works that take a first step to real large-scale 3D pre-training, as the 2D community does.

\vspace{-2mm}
\section*{Acknowledgements}
\vspace{-2mm}
This work is supported in part by the National Natural Science Foundation of China (No. 62201484), HKU Startup Fund, and HKU Seed Fund for Basic Research.

{
\small
\bibliographystyle{ieee_fullname}
\bibliography{main}
}

\clearpage
\newpage
\appendix
\section*{Appendix}
\begin{figure*}[t!]\centering
\includegraphics[width=0.95\linewidth]{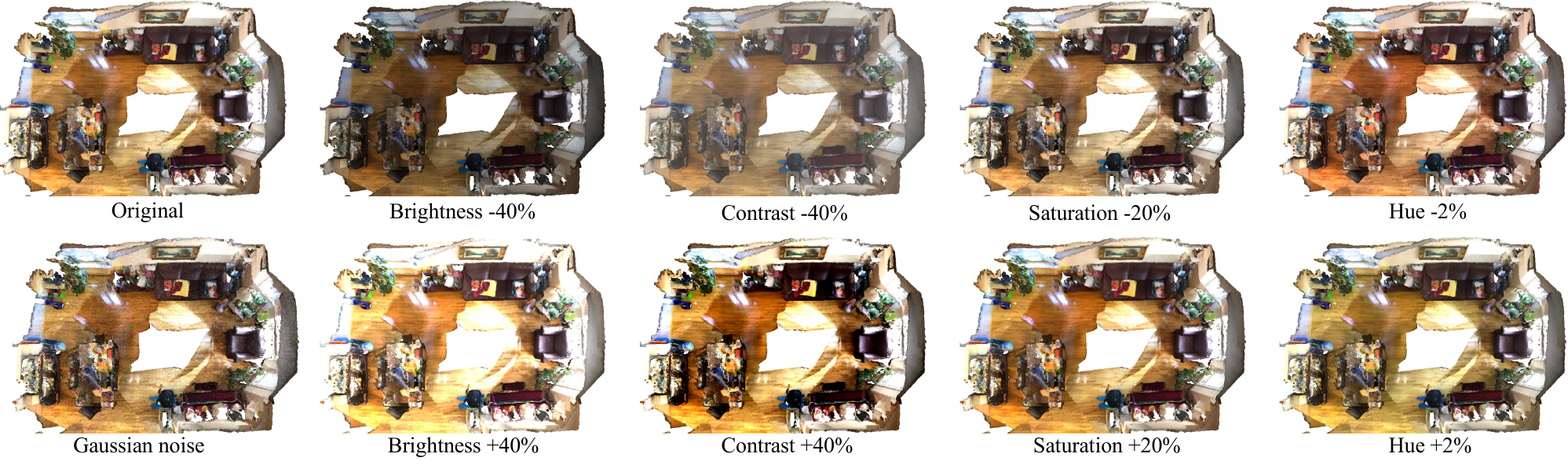}
    \caption{\textbf{Photometric augmentation.}}
    \label{fig:sup_augmentation}
\end{figure*}

\section{Implementation Details}
This section introduces the implementation details of our proposed \textit{Masked Scene Contrast} (MSC), which is crucial to making these novel designs work.
\subsection{Backbone Architecture}
We adopt \textit{SparseUNet}~\cite{choy20194d}, which is widely applied by previous works as ablation studies and result comparisons. \textit{SparseUNet} adopt a U-Net style architecture, and the config details follow previous works~\cite{xie2020pointcontrast, hou2021exploring, wu2022point}. The main config is available in \tabref{tab:sup_backbones}, and the name of the backbone is marked in \marktext{gray}{gray}.

\subsection{View Generation Pipeline.}
The specific constitution of our generation pipeline is concluded in \tabref{tab:sup_augmentation}. For a given point cloud input, we first dedicate two copies of the original point cloud for separated random view generation. Then we apply the augmentation sequence in \tabref{tab:sup_augmentation} to produce differentiated views of a single scene. The original coordinates (w/o rotation) are saved for both views, and both grid sampling and point matching are performed on this original coordinate system. Spatial augmentations, photometric augmentations, and sampling augmentations are marked in \marktext{green!5}{green}, \marktext{yellow!10}{yellow} and \marktext{blue!5}{blue}.

\mypara{Spatial augmentations.} We simulate different orientations of point cloud scenes by randomly rotating around the z-axis. Slight rotations around the x-axis and y-axis are also applied to simulate the unavoidable slope of the ground. Additional random flipping also adds geometric diversity to objects in the scenes and is thus also applied.

\begin{table}[t!]
\begin{minipage}{.48\textwidth}
    \centering
    \tablestyle{7pt}{1.08}
    \begin{tabular}{y{33mm}|x{39mm}}\toprule
Config &Value \\\cmidrule{1-2}
\rowcolor[HTML]{efefef}backbone & SparseUNet34 \\
patch embed depth &1 \\
patch embed channels &32 \\
patch embed kernel size &5 \\
encode depths &[2, 3, 4, 6] \\
encode channels &[32, 64, 128, 256] \\
encode kernel size &3 \\
decode depths &[2, 2, 2, 2] \\
decode channels &[256, 128, 64, 64] \\
decode kernel size &3 \\
pooling stride &[2, 2, 2, 2] \\
\bottomrule
\end{tabular}

    \vspace{-2mm}
    \caption{\textbf{Backbone setting.}}
    \vspace{2mm}
    \label{tab:sup_backbones}
\end{minipage} \\
\begin{minipage}{.48\textwidth}
    \centering
    \tablestyle{7pt}{1.08}
    \begin{tabular}{l|cc}\toprule
Augmentation &Value \\\midrule
\rowcolor{green!5} random rotate &angle=[-1, 1], axis=`z', p=1 \\
\rowcolor{green!5} random rotate &angle=[-1/64, 1/64], axis=`x', p=1 \\
\rowcolor{green!5} random rotate &angle=[-1/64, 1/64], axis=`y', p=1 \\
\rowcolor{green!5} random flip &p=0.5 \\
\rowcolor{yellow!5} random coord jitter &sigma=0.005, clip=0.02 \\
\rowcolor{yellow!5} random color brightness jitter &ratio=0.4, p=0.8 \\
\rowcolor{yellow!5} random color contrast jitter &ratio=0.4, p=0.8 \\
\rowcolor{yellow!5} random color saturation jitter &ratio=0.2, p=0.8 \\
\rowcolor{yellow!5} random color hue jitter &ratio=0.02, p=0.8 \\
\rowcolor{yellow!5} random color gaussian jitter &std=0.05, p=0.95 \\
\rowcolor{blue!5} grid sample &grid size=0.02 \\
\rowcolor{blue!5} random crop &ratio=0.6 \\
center shift &n/a \\
color normalze &n/a \\
\bottomrule
\end{tabular}

    \vspace{-2mm}
    \caption{\textbf{View generation pipeline.}}
    \vspace{2mm}
    \label{tab:sup_augmentation}
\end{minipage} \\
\begin{minipage}{.48\textwidth}
    \centering
    \tablestyle{5pt}{1.08}
    \begin{tabular}{y{33mm}|x{39mm}}
\toprule
Config &Value \\\midrule
optimizer &SGD \\
scheduler &cosine decay \\
learning rate &0.1 \\
weight decay &1e-4 \\
optimizer momentum &0.8 \\
batch size &32 \\
datasets &ScanNet, ArkitScene \\
warmup epochs &6 \\
epochs &600 \\
\bottomrule
\end{tabular}
    \vspace{-2mm}
    \caption{\textbf{Pre-training setting.}}
    \vspace{2mm}
    \label{tab:sup-pre-training}
\end{minipage} \\
\begin{minipage}{.48\textwidth}
    \centering
    \tablestyle{5pt}{1.08}
    \begin{tabular}{y{33mm}|x{39mm}}
\toprule
Config  &Value \\\midrule
optimizer &SGD \\
scheduler &cosine decay \\
learning rate &0.05 \\
weight decay &1e-4 \\
optimizer momentum &0.9 \\
batch size &48 \\
warmup epochs &40 \\
epochs &800 \\
\bottomrule
\end{tabular}
    \vspace{-2mm}
    \caption{\textbf{Fine-tuning setting.}}
    \label{tab:sup-fine-tuning}
\end{minipage}
\end{table}

\mypara{Photometric augmentations.}
Our photometric augmentations contain brightness, contrast, saturation, and hue adjusting from 2D images to 3D point clouds. These augmentations enhance the chromatic augmentations scheme introduced by Choy \etal~\cite{choy20194d} three years ago, and we hope these advanced photometric augmentations can also benefit future works. As for the augmentation parameters, we follow BYOL~\cite{grill2020bootstrap}, a reputed unsupervised representation learning framework for 2D images. We shrink the boundary of hue adjustment since the hue diversity of 3D indoor scenes is limited compared with image datasets. These stochastic photometric augmentations can effectively simulate diverse light conditions. A visualization of these augmentations is available in \figref{fig:sup_augmentation}.

\mypara{Sampling augmentations.} Grid sampling is a necessary process that both reduces point redundancy and increases data diversity. Combined with random rotation, the grid sampling is applied to different grids and points from the original point cloud, which adds to the data diversity. Further, random cropping is also applied to simulate the occlusion relationship and enforce the model to differentiate the visible region of contrastive views, which is also an important component.

\subsection{Training Setting.}
\mypara{Pre-training.} The default setting is in \tabref{tab:sup-pre-training}. We only utilize ScanNet point cloud scene data for efficient pre-training. And we adopt both ScanNet~\cite{dai2017scannet} and ArkitScene~\cite{dehghan2021arkitscenes} for large-scale pretraining.

\mypara{Fine-tuning.} The default setting for fine-tuning on ScanNet semantic segmentation is in \tabref{tab:sup-fine-tuning}. It is worth noting that good fine-tuning results rely on higher batch size. And the conclusion holds for most of our experimented downstream tasks. We use the same setting proposed by CSC~\cite{hou2021exploring} and adopt 48 as the fine-tuning batch size for downstream tasks.

\end{document}